\documentclass[11pt]{article}

\usepackage{titlesec}

\titlespacing*{\section}{0pt}{1.0ex plus 0.3ex minus 0.2ex}{0.6ex}
\titlespacing*{\subsection}{0pt}{0.8ex plus 0.2ex minus 0.2ex}{0.4ex}
\titlespacing*{\subsubsection}{0pt}{0.6ex plus 0.2ex minus 0.1ex}{0.3ex}

\usepackage{enumitem}
\usepackage[final]{acl}
\usepackage{amsmath}
\usepackage{stfloats}
\usepackage{times}
\usepackage{latexsym}
\usepackage{tabularx}
\usepackage{booktabs}
\usepackage{multirow}
\usepackage{float}
\usepackage[dvipsnames]{xcolor}
\usepackage{tabularx}
\usepackage{booktabs}
\usepackage{multirow}
\usepackage[table]{xcolor} 

\usepackage{amsmath}
\usepackage{amsfonts}
\usepackage{amssymb}

\usepackage[T1]{fontenc}
\usepackage[utf8]{inputenc}

\usepackage{microtype}

\usepackage{inconsolata}

\usepackage{graphicx}

\title{PolarMind at SemEval-2026 Task 9: \\ Leveraging LaBSE with Progressive Curriculum Learning for Multicultural Polarization}

\author{
  \textbf{Sandeep Kumar}$^\dagger$ \hspace{0.5cm}
  \textbf{Mothish}$^\clubsuit$  \hspace{0.5cm}
  \textbf{Sachin Sundar}$^\clubsuit$
 \\  \\
  $^\dagger$Indian Institute of Technology, Kharagpur,\\ $^\clubsuit$Indian Institute of Technology, Madras \\
  \texttt{sandeepkumar.24@kgpian.iitkgp.ac.in}  \\
\texttt{\{cs24b033, mm24b013\}@smail.iitm.ac.in}
}

\begin{document}
\maketitle
\begin{abstract}

Detecting online polarization remains a \\ critical challenge, particularly in multilingual and multicultural contexts where intergroup hostility is prevalent.  The problem is particularly challenging due to the data scarcity for these tasks in the low-resource languages. Identifying such phenomena has become an active area of research and is addressed in SemEval-2026 Task 9: Multilingual, Multicultural Online Polarization Detection. To address this problem we propose an architecture that leverages LaBSE embeddings—an unconventional choice typically reserved for retrieval tasks—to obtain strong cross-lingual learning which enhances scores in low-resource language by a score up to 0.2 macro F1. 
Furthermore, we provide a comprehensive ablation study evaluating the performance of diverse encoder models in the Qwen model family within a retrieval-based prompting framework.\footnote{Our code will be soon available at \url{https://github.com/carrycurious/PolarMind}}

\end{abstract}

\section{Introduction and Related Work}

\indent In the contemporary digital landscape, social networks have emerged as a primary medium of information exchange and a source through which people belonging to diverse populations interact. However, it has become common to see inter-group hostility and antagonistic viewpoints expressed across various platforms.\\  \indent
Digital platforms often amplify extreme views, degrading the quality of online debate \citep{IANDOLI2021120924}. Because of this, identifying polarized speech is now a critical challenge for the research community. Polarized text can be broadly classified into 4 main categories: political, religious, racial/ethnic, and gender/sexual. The major challenge for building reliable systems that detect these is that the text can be polarized due to a single harsh word or even due to the overall opinion expressed in the comment. Polarization is subject to different ethnic groups and cultures \citep{kannen2025aestheticsculturalcompetencetexttoimage}, i.e., a completely unpolarised sentence in English when translated to Urdu might be polarized\citep{zafar2025investigating}. Consequently, the optimal method to approach this problem would be to design specialized systems for each language. However, most of the state-of-the-art models today exhibit large performance gaps between English and low-resource languages\citep{verma2025milumultitaskindiclanguage,Saji_2025}. Also, with a limited amount of training data, it becomes necessary for our systems to group languages together in-order to gain better overall performance through cross-lingual learning.
To address these challenges, this paper presents our approach for SemEval 2026 task 9:"Multilingual Text Classification Challenge"\citep{naseem-etal-2026-polar}. We participate in 2 subtasks each with 22 languages in each subtask \citep{naseem2026polarbenchmarkmultilingualmulticultural}.

\begin{itemize}[nosep, leftmargin=*]
  \item \textbf{Subtask 1:} Polarization detection (binary classification)
  \item \textbf{Subtask 2:} Polarization type detection with 5 classes 
  (Political, Racial, Gender, Ethnic, Other)
\end{itemize}
\vspace{-0.2em}

\noindent In this work, we propose a unique architecture using layer fusion and hybrid 
pooling (attention pooling + mean pooling) in LaBSE embeddings along with 
Proxy-Guided Curriculum Learning to get enhanced results. In a detailed 
ablation study, we compare RemBERT with and without continual pretraining, 
EuroBERT — a state-of-the-art model with more recent knowledge — and 
Qwen-2.5-14B using in-context learning with MMR-reranked sentence retrieval. 
For more details regarding our model architecture and the proxy-guided 
curriculum learning strategy, refer to Section \ref{sec:methodology}.

\paragraph{Contributions}
\begin{itemize}
    \item LaBSE-based architecture with weighted layer aggregation and hybrid pooling for cross-lingual polarization detection.
    \item Proxy-guided curriculum learning to address multilingual data imbalance.
    \item Comparative evaluation of encoder models and LLMs (Qwen-2.5).
    \item Empirical analysis of LLM and phoneme prompting limitations in low-resource settings.
\end{itemize}

\section{System Architecture and Ablations}
\label{sec:methodology}
This section details our primary system submitted for the final shared task evaluation: a customized LaBSE architecture. To justify our design choices, we also present an ablation study comparing this core system against alternative models.

\vspace{-0.2em}
\subsection{Core Architecture: LaBSE with Hybrid Pooling}
\indent We employ LaBSE \citep{feng2022languageagnosticbertsentenceembedding}, a bi-encoder architecture designed to map multilingual sentences with equivalent semantics into a shared embedding space. 
Unlike standard multilingual encoders like mBERT or XLM-R, LaBSE is fundamentally optimized to map semantically similar cross-lingual sentences to the exact same vector space. We hypothesize that this strict alignment is uniquely beneficial for multicultural polarization, allowing the model to project scarce low-resource polarized phrases into the robust semantic neighborhoods established by high-resource languages.
The main components of our architecture are as follows:
\vspace{-0.6em}
\paragraph{Weighted Layer Aggregation:} This model aggregates the final four layers of the models with learnable parameters ($w_1, w_2, w_3, w_4$), where ($w_1, w_2, w_3$) are initialized to be much smaller ($\sim 0.2w_1$).
\vspace{-0.8em}
\begin{equation}
    H_{final} = \sum_{i=1}^{4} w_i \cdot L_{n-4+i}
\end{equation}
\paragraph{Hybrid pooling:} We propose a hybrid pooling mechanism that computes a linear combination of global mean pooling and additive-attention \\ \citep{bahdanau2016neuralmachinetranslationjointly} pooling. This approach is designed to prioritize and highlight the most significant tokens.

% --- Mean Pooling ---

The global semantic context is captured by calculating the arithmetic mean of the hidden states:
\begin{equation}
    \mathbf{e}_{\text{mean}} = \frac{1}{T} \sum_{i=1}^{T} \mathbf{h}_i
\end{equation}
where $T$ denotes the sequence length and $\mathbf{h}_i$ represents the $i$-th hidden vector.

% --- Attention Pooling ---

To emphasize high-impact tokens, we employ an additive attention mechanism:
\begin{align}
    u_i &= \tanh(\mathbf{W} \mathbf{h}_i + \mathbf{b}) \\
    \alpha_i &= \frac{\exp(\mathbf{w}^\top u_i)}{\sum_{j=1}^{T} \exp(\mathbf{w}^\top u_j)} \\
    \mathbf{e}_{\text{attn}} &= \sum_{i=1}^{T} \alpha_i \mathbf{h}_i
\end{align}
where $\mathbf{W} \in \mathbb{R}^{k \times d}$ and $\mathbf{b}$ are learnable parameters, and $\mathbf{w}^\top$ is the context vector.

% --- Hybrid Pooling ---

The final representation is a weighted sum of the global and salient features, modulated by the hyperparameter $\lambda$:
\begin{equation}
    \mathbf{e}_{\text{hybrid}} = \mathbf{e}_{\text{mean}} + \lambda \mathbf{e}_{\text{attn}}
\end{equation}
The mean pooling component ($e_{\text{mean}}$) stabilizes the representation 
by capturing overall semantic context, while the additive attention mechanism 
($e_{\text{attn}}$) emphasizes high-impact tokens relevant to classification. 
In Subtask 2, each classification head maintains an independent attention 
mechanism, ensuring that token importance for one polarization type 
(e.g., gender) does not conflate with another (e.g., ethnic). The final 
representation $\mathbf{e}_{\text{hybrid}}$ is used for classification.

% --- NEW SUBSECTION 2.2 BOUNDARY ---
\subsection{Ablation Study: Alternative Models}
To evaluate the effectiveness of our core LaBSE architecture, we explored alternative models spanning both encoder-only and large language model (LLM) paradigms.

\paragraph{ii) RemBERT with continual pretraining:}
\indent To further extend our study, we leverage RemBERT \citep{chung2020rethinkingembeddingcouplingpretrained}, a high-capacity multilingual encoder, for the classification task. We perform continual pre-training on our dataset to adapt the model and make it aware of the recent events, as the model was released in 2020. For continual pretraining, we create a corpus using all the data in Subtask 3 (which we did not take part in) and use this for pretraining.
\paragraph{iii) EuroBERT-210m:}
\indent EuroBERT \citep{boizard2025eurobertscalingmultilingualencoders} is a state-of-the-art encoder model, optimized for English and european languages. EuroBERT (2024) has a more recent knowledge cut-off and hence has seen more information based on recent global events. We use this model to analyze the effect of recent knowledge.

\paragraph{iv) \textsc{Qwen-2.5} with retrieval using phonemes}: Recent advancements have shown that modern LLMs with optimized prompting, exhibit comparable performance with traditional encoder models in classification tasks\citep{wang2024chatgptgoodsentimentanalyzer}.
We use few-shot prompting \citep{brown2020languagemodelsfewshotlearners} with the following strategies:

\begin{figure*}[t] % Use [t] to place it at the top of the page
    \centering
    \includegraphics[width=1.0\textwidth]{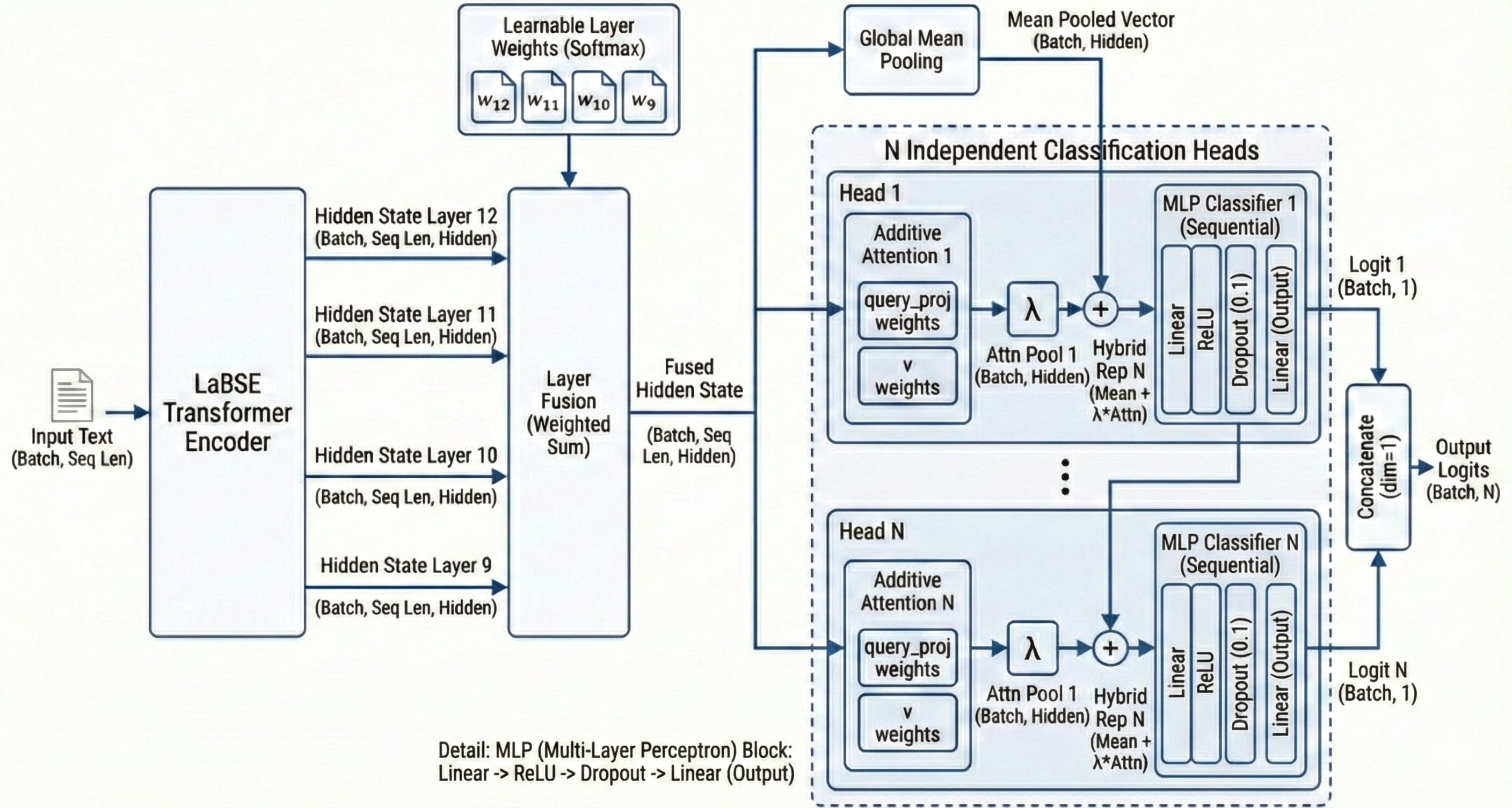} 
    \caption{Proposed Hybrid Contextual Pooling architecture.}
    \label{fig:hybrid_pooling}
\end{figure*}

\definecolor{tablegray}{gray}{0.93}

% Define a professional academic maroon
\definecolor{bestmaroon}{RGB}{128, 0, 0}

\newcommand{\test}[1]{{\color{MidnightBlue}\textsubscript{[#1]}}}

\begin{table*}[b]
\centering
\scriptsize 
\setlength{\tabcolsep}{3pt} 
\begin{tabularx}{\textwidth}{@{} l l l *{11}{X} @{}}
\toprule
\rowcolor{tablegray} % Shading the header row mildly
\textbf{Model} & \textbf{Strategy} & \textbf{Pooling} & \textbf{amh} & \textbf{arb} & \textbf{ben} & \textbf{deu} & \textbf{eng} & \textbf{fas} & \textbf{hau} & \textbf{hin} & \textbf{ita} & \textbf{khm} & \textbf{mya} \\
\midrule
\multirow{2}{*}{LaBSE} & mixed & ours & 0.723\test{.77} & 0.783\test{.81} & \textbf{0.818\test{.82}} & \textbf{0.735\test{.69}} & 0.804\test{.80} & \textbf{0.883\test{.81}} & \textbf{0.838\test{.82}} & \textbf{0.843\test{.77}} & 0.647\test{.64} & 0.697\test{.71} & \textbf{0.923\test{.86}} \\
 & separate & ours & \textbf{0.753} & \textbf{0.812} & 0.801 & 0.627 & 0.798 & 0.765 & 0.798 & 0.835 & 0.618 & \textbf{0.719} & 0.857 \\
\midrule
Vanilla RemBERT & mixed & Cls & 0.698 & 0.792 & 0.796 & 0.709 & 0.741 & 0.821 & 0.693 & 0.842 & \textbf{0.663} & 0.652 & 0.852 \\
\rowcolor{tablegray} % Striping this row to separate it visually
RemBERT (cont.) & mixed & Cls & 0.679 & 0.784 & 0.811 & 0.674 & 0.768 & 0.834 & 0.710 & 0.838 & 0.661 & 0.638 & 0.876 \\
EuroBERT & mixed & Cls & --- & --- & --- & 0.729 & \textbf{0.818} & --- & --- & --- & 0.586 & --- & --- \\
\midrule
\multirow{3}{*}{Qwen-2.5-Instruct-14b} & 0 shots & text & 0.598 & 0.765 & 0.732 & 0.728 & 0.741 & 0.649 & 0.582 & 0.643 & 0.485 & 0.127 & 0.548 \\
 & 7 shots & text & 0.685 & 0.740 & 0.698 & 0.687 & 0.764 & 0.698 & 0.601 & 0.689 & 0.517 & 0.527 & 0.765 \\
 & 7 shots & text+ipa & 0.624 & 0.756 & 0.687 & 0.719 & --- & 0.746 & 0.724 & 0.658 & 0.565 & --- & 0.793 \\

\bottomrule
\end{tabularx}

\label{tab:full_results}
\end{table*}

\vspace{-0.6em}
\paragraph{ Multilingual Sentence Retrieval}
We use LaBSE as a retriever to find the most(top k) semantically similar sentences from our training data for a given input sentence. This allows the model to see and understand how similar polarization manifests across different languages and cultural contexts before making a final prediction.

% ... in your document body ...
\vspace{-0.6em}
\paragraph{ Maximal Marginal Relevance (MMR) for Diversity}
To ensure that retrieved examples are both relevant and diverse, we implement the Maximal Marginal Relevance (MMR) algorithm \citep{10.1145/290941.291025}.
This algorithm ensures that prompt redundancy,i.e, makes sure that prompts are similar to the given query $Q$ but yet diverse.
 For a given query $Q$, MMR iteratively selects a sentence $D_i$ from candidate set $R$ as follows:

\begin{equation} \label{eq:mmr}
\footnotesize
\begin{aligned}
\text{MMR} = \arg \max_{D_i \in R \setminus S} \bigl[ \alpha \cdot \text{Sim}_1(D_i, Q) \\
- (1 - \alpha) \cdot \max_{D_j \in S} \text{Sim}_2(D_i, D_j) \bigr]
\end{aligned}
\end{equation}

where $\text{Sim}_1(D_i, Q)$ is the cosine similarity between candidate 
$D_i$ and query $Q$ (relevance), $\max \text{Sim}_2(D_i, D_j)$ is the 
maximum similarity to already selected sentences $S$ (diversity), and 
$\alpha{=}0.7$ balances the two.
\paragraph{ Few-Shot Prompting with Qwen-2.5}
The top $k$ diverse sentences selected via MMR are formatted into a structured prompt as shown in the Appendix \ref{sec:prompts}. Using this retrieved context, Qwen-2.5-14B \citep{qwen2025qwen25technicalreport} performs in-context learning to categorize the target input, leveraging the diverse examples to better capture the nuances of the classification task.

\begin{table*}[t]
\centering
\scriptsize 
\setlength{\tabcolsep}{3pt} 
\begin{tabularx}{\textwidth}{@{} l l l *{11}{X} @{}}
\toprule
\rowcolor{tablegray} % Shading the header row mildly
\textbf{Model} & \textbf{Strategy} & \textbf{Pooling} & \textbf{nep} & \textbf{ori} & \textbf{pan} & \textbf{pol} & \textbf{rus} & \textbf{spa} & \textbf{swa} & \textbf{tel} & \textbf{tur} & \textbf{urd} & \textbf{zho} \\
\midrule
\multirow{2}{*}{LaBSE} & mixed & ours & \textbf{0.886\test{.90}} & 0.771\test{.79} & \textbf{0.819\test{.76}} & 0.768\test{.77} & \textbf{0.778\test{.75}} & 0.685\test{.75} & \textbf{0.798\test{.79}} & 0.881\test{.87} & 0.799\test{.76} & \textbf{0.727\test{.76}} & \textbf{0.878\test{.85}} \\
 & separate & ours & 0.859 & \textbf{0.791} & 0.789 & 0.758 & 0.768 & 0.659 & 0.754 & 0.885 & 0.776 & 0.683 & 0.874 \\
\midrule
Vanilla RemBERT & mixed & Cls & 0.850 & 0.701 & 0.810 & \textbf{0.804} & 0.775 & \textbf{0.696} & 0.778 & 0.908 & 0.790 & 0.693 & 0.867 \\
\rowcolor{tablegray} % Striping this row to separate it visually
RemBERT (cont.) & mixed & Cls & 0.837 & 0.728 & 0.803 & 0.798 & 0.757 & 0.672 & 0.758 & \textbf{0.912} & \textbf{0.804} & 0.723 & 0.852 \\
EuroBERT & mixed & Cls & --- & --- & --- & 0.750 & --- & 0.692 & --- & --- & --- & --- & --- \\
\midrule
\multirow{3}{*}{Qwen-2.5-14b} & 0 shots & text & 0.736 & 0.685 & 0.578 & 0.717 & 0.636 & 0.643 & 0.569 & 0.576 & 0.751 & 0.659 & 0.794 \\
 & 7 shots & text & 0.725 & 0.692 & 0.767 & 0.735 & 0.578 & 0.623 & 0.690 & 0.713 & 0.758 & 0.689 & 0.864 \\
 & 7 shots & text+ipa & 0.745 & --- & --- & 0.718 & 0.658 & 0.672 & 0.728 & 0.735 & 0.748 & 0.719 & 0.856 \\
\bottomrule
\end{tabularx}

\caption{Macro F1 scores for Subtask 1 (Dev and Official Test sets). 
LaBSE subscripts denote official Test results. For Qwen, ``Strategy'' 
= shot count, ``Pooling'' = prompting technique. \textbf{Bold} = best per language.}
\label{tab:full_results}
\vspace{-0.5em}
\end{table*}

\paragraph{Phonemes based prompting}
\citet{Nguyen_2025} has shown that introducing phoneme information to 
LLMs along with raw scripts (\textit{text+ipa}) improves performance in 
various downstream tasks. Following this approach, we utilize the 
\textit{Epitran} \citep{mortensen-etal-2018-epitran} library to convert 
text from each language into phonemes for further analysis.

 \begin{figure}[H] % Use [H] with \usepackage{float} to stay "HERE"
    \centering
    \includegraphics[width=\columnwidth]{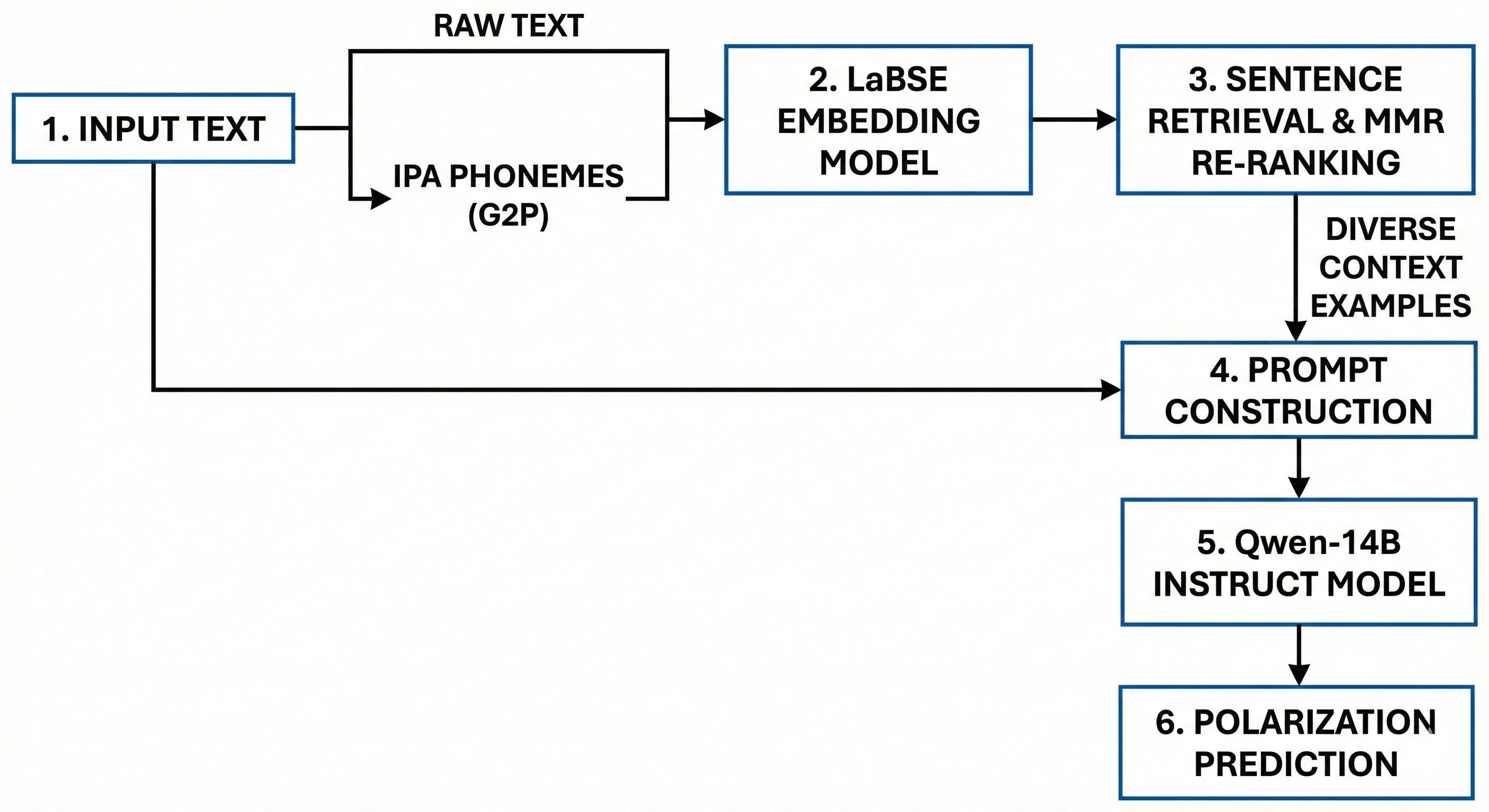} 
    \caption{Pipeline that integrates phonemes}
    \label{fig:phonemes}
    \vspace{-1.5em}
\end{figure}

% Define a very mild professional gray for striping

\definecolor{tablegray}{gray}{0.93}

\
\begin{table*}[h!]
\centering
\scriptsize 
\setlength{\tabcolsep}{3pt} 
\begin{tabularx}{\textwidth}{@{} l l l *{11}{X} @{}}
\toprule
\rowcolor{tablegray} % Shading the header row mildly
\textbf{Model} & \textbf{Strategy} & \textbf{Pooling} & \textbf{amh} & \textbf{arb} & \textbf{ben} & \textbf{deu} & \textbf{eng} & \textbf{fas} & \textbf{hau} & \textbf{hin} & \textbf{ita} & \textbf{khm} & \textbf{mya} \\
\midrule
LaBSE & mixed & ours & \textbf{0.469\test{.63}} & \textbf{0.564\test{.59}} & 0.350\test{.34} & \textbf{0.526\test{.55}} & 0.430\test{.48} & 0.551\test{.58} & 0.234\test{.41} & \textbf{0.755\test{.77}} & 0.371\test{.31} & \textbf{0.645\test{.61}} & \textbf{0.545\test{.70}} \\
\midrule
\multirow{2}{*}{Qwen-2.5-14b} 
 & 7 shots & text  & 0.423 & 0.498 & 0.442 & 0.501 & 0.497 & \textbf{0.569} & \textbf{0.477} & 0.547 & 0.537 & 0.574 & 0.457\\
 & 7 shots & text+ipa & 0.417 & 0.512 & \textbf{0.521} & 0.513 & \textbf{0.499} & 0.562 & 0.472 & 0.578 & \textbf{0.541} & 0.602 & 0.468 \\
\bottomrule
\end{tabularx}
\label{tab:full2}
\end{table*}

\begin{table*}[]
\centering
\scriptsize 
\setlength{\tabcolsep}{3pt} 
\begin{tabularx}{\textwidth}{@{} l l l *{11}{X} @{}}
\toprule
\rowcolor{tablegray} % Shading the header row mildly
\textbf{Model} & \textbf{Strategy} & \textbf{Pooling} & \textbf{nep} & \textbf{ori} & \textbf{pan} & \textbf{pol} & \textbf{rus} & \textbf{spa} & \textbf{swa} & \textbf{tel} & \textbf{tur} & \textbf{urd} & \textbf{zho} \\
\midrule
LaBSE & mixed & ours & \textbf{0.758\test{.77}} & 0.501\test{.51} & 0.397\test{.42} & 0.534\test{.51} & 0.469\test{.49} & \textbf{0.607\test{.63}} & 0.455\test{.47} & 0.463\test{.44} & \textbf{0.595\test{.56}} & \textbf{0.754\test{.76}} & 0.717\test{.73} \\
\midrule
\multirow{2}{*}{Qwen-2.5-14b} 
 & 7 shots & text & 0.578 & 0.593 & 0.565 & \textbf{0.563} & \textbf{0.553} & 0.552 & \textbf{0.548} & \textbf{0.540} & 0.539 & 0.535 & 0.511 \\
 & 7 shots & text+ipa & 0.592 & \textbf{0.643} & \textbf{0.609} & 0.524 & 0.513 & 0.568 & 0.531 & 0.517 & 0.483 & 0.562 & \textbf{0.793} \\
\bottomrule
\end{tabularx}
\caption{Macro F1 scores for Subtask 2 on Development and Official Test set. Bolding indicates the highest value per column. Same guidelines as Table ~\ref{tab:full_results}}
\label{tab:full1}
\vspace{-0.7em}
\end{table*}

\section{Experimental setup}
\subsection{Dataset}
For both evaluation and training of our data, we use the dataset provided by organisers of Semeval task 9 Organizers \citep{naseem2026polarbenchmarkmultilingualmulticultural}. For all subsequent analyses, we report performance metrics as detailed in Table~\ref{tab:full_results} on development set only.
\paragraph{Proxy guided Curriculum Learning.} We employ a curriculum learning strategy for training all encoder models described in Section~\ref{sec:methodology}.To implement curriculum learning, we first categorize the dataset by difficulty using XLM-RoBERTa-base \citep{conneau-etal-2020-unsupervised}, a smaller model for our purpose. We then rank the entire data based on the following equation:
    $$R(x) = \alpha \cdot \text{P}(y|x) + \beta \cdot \frac{1}{\text{len}(x)}$$
where $\text{P}(y|x)$ is the proxy confidence and $\beta/\text{len}(x)$ 
promotes lower-fertility high-resource languages to higher ranks, 
partitioning data into \textit{easy}, \textit{medium}, and \textit{hard} 
training groups.

\subsection{Training setup}
\paragraph{Encoder models} Encoder models are trained with the same 
strategy for both subtasks, differing only in the number of classification 
heads. Each model undergoes four epochs following the curriculum: easy 
samples first, then easy+medium, then medium+hard, and finally the full 
dataset. We use AdamW ($lr=2e{-5}$, weight decay $0.01$) with a Cosine 
Annealing Scheduler \citep{loshchilov2016sgdr}, trained on a single NVIDIA 
L4 GPU using HuggingFace \texttt{transformers} \citep{transformers}. Full 
hyperparameters are in Appendix \ref{sec:appendix_hyperparameters}.
\vspace{-1.0em}
\paragraph{Language-Specific Decision Calibration}
For the multi-label task, we apply sigmoid activation and calibrate 
language- and class-specific thresholds $t_{\ell,c}$ on a 20\% training 
split to maximize Macro-F1:
\begin{equation}
t_{\ell,c}^{*} = \arg\max_{t} \text{MacroF1}_{\ell,c}(t)
\end{equation}

\section{Results}
\vspace{-0.5em}
Overall, we evaluate our proposed methods and their variants on the development set to identify the optimal model configuration for each language (indicated in bold in Table~\ref{tab:full_results} and Table~\ref{tab:full1}). Our performance in low-resource languages is highly competitive on the official leaderboard; we achieved top-10 rankings in five languages in Subtask 1. In Subtask 2, we secured top-5 placements for five languages and top-10 rankings for eleven languages globally.

\begin{figure*}[b] % Use [t] to place it at the top of the page
    \centering
    \includegraphics[width=1.0\textwidth]{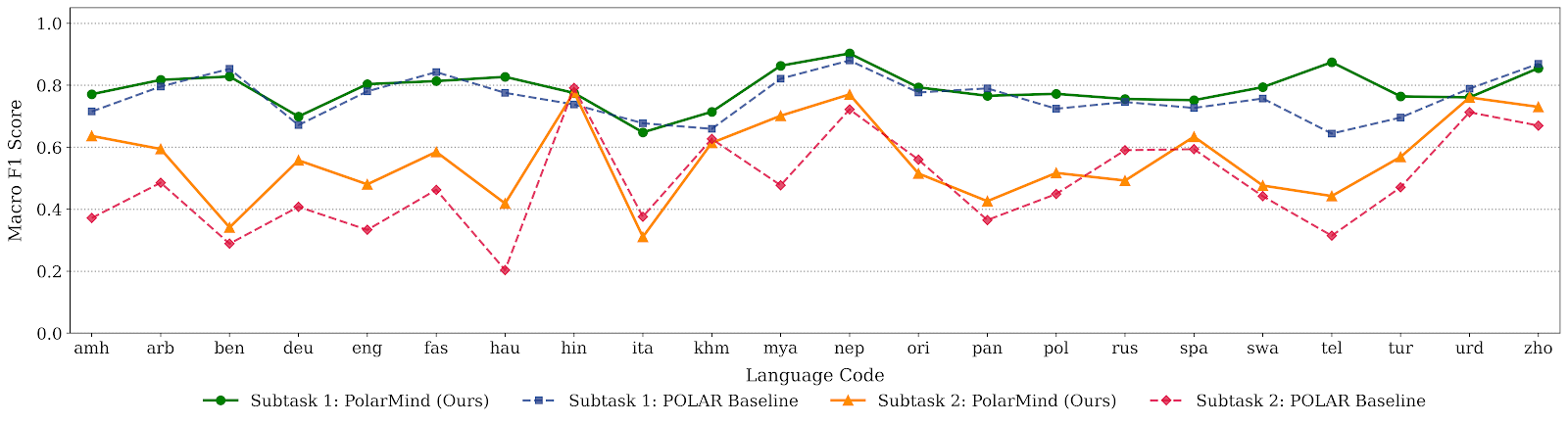} 
    \caption{Comparison with POLAR baseline}
    \label{fig:hybr}
\end{figure*}
\subsection{Subtask 1}
\paragraph{Development set} 
Our proposed LaBSE-based architecture with hybrid context pooling consistently outperformed the official \texttt{POLAR baseline} (fine-tuned LaBSE-base) by an average of \textbf{0.04 Macro F1 points} across all 22 languages. On the official SemEval-2026 leaderboard, our system demonstrated strong performance across low-resource languages, particularly in low-resource scripts. For instance, in Amharic (\texttt{amh}), we achieved a score of \textbf{0.771}, a \textbf{+5.6\% improvement} over the baseline. \\
\vspace{-0.1em}
While \textsc{EuroBERT} (210M) performed at par in specific European contexts—surpassing our model in English and Italian by approximately 0.01 Macro F1—this is likely due to its more recent knowledge cut-off tailored. However, the marginal gains (\textbf{+0.005 Macro F1}) observed from continual pre-training (\textsc{RemBERT-cont}) suggest that our hybrid pooling strategy provides a more significant performance boost than model scaling alone. For the decoder model (Qwen-2.5), performance is not competitive with encoder models, likely due to tokenization mismatches across scripts and weaker cross-lingual alignment compared to LaBSE. Phonetic information does not significantly improve performance. We hypothesize that the phonetic layer offers redundant information for this specific task, as the model’s internal representations of raw text already encompass the necessary cultural nuances, while the IPA transformation potentially filters out emoji usage and script-specific markers that carry significant polarized weight.

\subsection{Subtask 2}
\paragraph{Development set}
As shown in Table~\ref{tab:full1}, we focus on our LaBSE-based model as it consistently outperforms other baselines (RemBERT and EuroBERT). Language-specific threshold calibration improved the average Macro-F1 on the development set from 0.5216 to 0.5441, an absolute gain of +2.25\%. These gains are most evident in languages with skewed class distributions, emphasizing the sensitivity of multi-label polarization detection to threshold selection.

Interestingly, Qwen (7-shot) performs on par with our encoder model. While the encoder struggles to generalize sparse patterns (e.g., gender or ethnicity) from limited samples, Qwen leverages its prior knowledge via few-shot prompting. Finally, while our phoneme-based approach yields minor improvements, the current results do not provide a clear pattern or conclusive evidence that modern LLMs benefit from phonetic prompting for this task.

\vspace{-0.5em}
\section{Conclusion}
\vspace{-0.5em}
We evaluate our proposed methods and their variants on the development set to identify the optimal model configuration for each language (indicated in bold in Table~\ref{tab:full_results} and Table~\ref{tab:full1}). Our performance in low-resource languages is highly competitive on the official leaderboard; we achieved top-10 rankings for five languages in Subtask 1. In Subtask 2, we secured top-5 placements for five languages and top-10 rankings for eleven languages globally.

\bibliography{custom}

\appendix

\section{Hyperparameters}
\label{sec:appendix_hyperparameters}

All models were trained on an NVIDIA L4 GPU. The encoder backbones were \textbf{LaBSE} and \textbf{RemBERT}. The \textbf{Qwen} model was used only for evaluation (no fine-tuning).

\vspace{0.2cm}
\noindent\textbf{Training Setup} 

\begin{center}
\small % Makes the table smaller
\renewcommand{\arraystretch}{1.2} % Adds space between rows
\setlength{\tabcolsep}{10pt} % Adds space between columns
\begin{tabular}{ll}
\hline
\textbf{Hyperparameter} & \textbf{Value} \\ \hline
Optimizer & AdamW ($\epsilon=1 \times 10^{-5}$) \\
Batch Size & 32 \\
Scheduler & Cosine decay \\
Total Steps & $\text{len(dataloader)} \times 4$ \\ \hline
\end{tabular}
\end{center}
\vspace{0.2cm}

Subtask 1 used Cross-Entropy loss, while Subtask 2 used weighted inverse-frequency Cross-Entropy.

Fusion pooling used $\lambda=0.2$ (Subtask 1) and $\lambda=0.5$ (Subtask 2).  
For LaBSE layer weighting: $W_1=W_2=W_3=0.2$, $W_4=1$.

\vspace{0.2cm}

\begin{table}[H]
\centering
\small % Sets a clean base font size
\renewcommand{\arraystretch}{1.25} % Adds vertical space so rows don't overlap
\resizebox{\columnwidth}{!}{% Scaled to fit column width perfectly
\begin{tabular}{lcc}
\toprule
\textbf{Parameter} & \textbf{LaBSE} & \textbf{RemBERT} \\
\midrule
\multicolumn{3}{l}{\textit{Architecture Config}} \\
Max Token Length & 512 & 512 \\
Hidden Size & 768 & 1152 \\
Encoder Layers & 12 & 32 \\
Attention Heads & 12 & 18 \\
Dropout & 0.1 & 0.1 \\
\midrule
\multicolumn{3}{l}{\textit{General Hyperparameters}} \\
Batch Size & 32 & 32 \\
Epochs & 4 (Curriculum) & 4 (Curriculum) \\
AdamW $\epsilon$ & $1 \times 10^{-5}$ & $1 \times 10^{-5}$ \\
Weight Decay & 0.01 & 0.01 \\
Warm-up Ratio & 0.1 & 0.1 \\
Gradient Clipping & 1.0 & 1.0 \\
\midrule
\multicolumn{3}{l}{\textit{Differential Learning Rates}} \\
Embedding + First 3 Layers & $1 \times 10^{-6}$ & -- \\
Remaining Encoder Layers & $1 \times 10^{-5}$ & $1 \times 10^{-5}$ \\
Self-Attention + Head & $5 \times 10^{-5}$ & $5 \times 10^{-5}$ \\
\bottomrule
\end{tabular}
}
\caption{Model architecture and training hyperparameters for both encoder backbones.}
\label{tab:hyperparams}
\end{table}

\section{Prompts}
\label{sec:prompts}
To illustrate our methodology, we provide representative 1-shot examples of the prompt templates used for both subtasks. While our experimental setup utilized 7-shot prompting, these examples highlight the structural integration of the text and its corresponding phonetic (IPA) transcription.

\subsection{Subtask 1: Polarization Detection}
\label{subsec:prompt_s1}

\begin{center}
\setlength{\fboxsep}{8pt}
\fbox{
\begin{minipage}{0.93\linewidth}
\footnotesize
\textbf{\texttt{System:}} You are a strict classifier. Categorize whether the input text is polarized or not. \\

\textbf{\texttt{User:}} \\
\textbf{Text:} Hawa Ni moja ya Watu wabaya sana Mufirisi tunaowaabudu hapa Tz WAHINDI WAARABU WACHINA Ameniambia Mzee Hussen hapa Ila AnyWay \\
\textbf{IPA:} hawa ni moa ja watu waaja sana mufiisi tunaowaauu hapa tz wahindi waaau watina ameniambia mzee hussen hapa ila awaj \\
\textbf{Is the text Polarized?} \\

\textbf{\texttt{Assistant:}} YES \\

\textbf{\texttt{User:}} \\
\textbf{Text:} hao ni wakikuyu wanyonge sana sisi wengine hatukupigia kura huko \\
\textbf{IPA:} hao ni wakikuju waone sana sisi wenine hatukupiia kua huko \\
\textbf{Is the text Polarized?} \\

\textbf{\texttt{Assistant:}} 
\end{minipage}
}
\end{center}

\vspace{0.5em}
\subsection{Subtask 2: Polarization Type Detection}
\label{subsec:prompt_s2}

\begin{center}
\setlength{\fboxsep}{8pt}
\fbox{
\begin{minipage}{0.93\linewidth}
\footnotesize
\textbf{\texttt{System:}} You are a strict classifier. Identify the categories of polarization present in the text. \\

\textbf{\texttt{User:}} \\
\textbf{Text:} Hawa Ni moja ya Watu wabaya sana Mufirisi tunaowaabudu hapa Tz WAHINDI WAARABU WACHINA Ameniambia Mzee Hussen hapa Ila AnyWay \\
\textbf{IPA:} hawa ni moa ja watu waaja sana mufiisi tunaowaauu hapa tz wahindi waaau watina ameniambia mzee hussen hapa ila awaj \\

\textbf{\texttt{Assistant:}} \\
\textbf{Labels (political, racial, religious, gender, other):} NO, YES, NO, NO, NO \\

\textbf{\texttt{User:}} \\
\textbf{Text:} hao ni wakikuyu wanyonge sana sisi wengine hatukupigia kura huko \\
\textbf{IPA:} hao ni wakikuju waone sana sisi wenine hatukupiia kua huko \\

\textbf{\texttt{Assistant:}} \\
\textbf{Labels (political, racial, religious, gender, other):}
\end{minipage}
}
\end{center}

\end{document}